\documentclass[letterpaper, 10 pt, conference]{ieeeconf}  %

\IEEEoverridecommandlockouts                              %

\usepackage{graphicx} %
\usepackage{times}
\usepackage{amsmath,amssymb}  %
\usepackage{verbatim}
\usepackage{mathtools}
\usepackage{algpseudocode}
\usepackage{tabularx} 
\usepackage{multirow}
\usepackage{soul, color} 
\usepackage{float}
\usepackage[table,x11names]{xcolor}
\usepackage{url}
\usepackage{hyperref}
\usepackage{cite}

\usepackage[export]{adjustbox}
\usepackage{rotating}
\usepackage{booktabs, makecell, tabularx}
\usepackage{algorithm}
\usepackage{hhline}

\usepackage{xcolor}

\graphicspath{ {./images/} }

\newcommand{\norm}[1]{\left\lVert#1\right\rVert}

\algnewcommand{\OR}{\algorithmicor}
\algnewcommand{\AND}{\algorithmicand}

\makeatletter
\def\endthebibliography{%
  \def\@noitemerr{\@latex@warning{Empty `thebibliography' environment}}%
  \endlist
}
\makeatother

\title{\LARGE \bf
\centering Robust Vehicle Localization and Tracking in Rain using Street Maps
}

\author{Yu Xiang Tan, Malika Meghjani
     \thanks{Yu Xiang Tan and Malika Meghjani are with Singapore University of Technology and Design, Singapore. {\tt\small yuxiang\_tan@mymail.sutd.edu.sg,  malika\_meghjani@sutd.edu.sg}} %
}

\begin{document}
\maketitle
\thispagestyle{empty}
\pagestyle{empty}

\begin{abstract}
GPS-based vehicle localization and tracking suffers from unstable positional information commonly experienced in tunnel segments and in dense urban areas. 
Also, both Visual Odometry (VO) and Visual Inertial Odometry (VIO) are susceptible to adverse weather conditions that causes occlusions or blur on the visual input. 
In this paper, we propose a novel approach for vehicle localization that uses street network based map information to correct drifting odometry estimates and intermittent GPS measurements especially, in adversarial scenarios such as driving in rain and tunnels. Specifically, our approach is a flexible fusion algorithm that integrates intermittent GPS, drifting IMU and VO estimates together with 2D map information for robust vehicle localization and tracking. We refer to our approach as Map-Fusion. 
We robustly evaluate our proposed approach on four geographically diverse datasets from different countries ranging across clear and rain weather conditions. These datasets also include challenging visual segments in tunnels and underpasses. We show that with the integration of the map information, our Map-Fusion algorithm reduces the error of the state-of-the-art VO and VIO approaches across all datasets. 
We also validate our proposed algorithm in a real-world environment and in real-time on a hardware constrained mobile robot. Map-Fusion achieved $2.46m$ error in clear weather and $6.05m$ error in rain weather for a $150m$ route.

\end{abstract}

\section{Introduction}
Robust localization is required for real-time fleet monitoring or traffic monitoring to operate seamlessly in intermittent GPS and all weather conditions. It is important to have accurate and robust vehicle tracking system as missing traffic information could negatively impact downstream tasks such as traffic management in smart cities or travel time prediction. 

Consider a scenario where a vehicle needs to be tracked in an adversarial setting of heavy rain and multiple tunnel segments.
The tunnel segments hinder GPS information while the raindrops remain on the camera lenses (even under shelter) compromising the visual input for Visual Odometry (VO) and Visual Inertial Odometry (VIO).
Specifically, the raindrops cause occlusions and lens flare creating additional visual artefacts \cite{garg_vision_2007}. Both of which causes VO and VIO algorithms to drift or delocalize \cite{tan_evaluating_2023}.
In such scenarios, only the sensors which are not affected by external factors could be relied. These include, the Inertial Measurement Unit (IMU). However, IMU drifts easily without correction especially in a long segment. This drift could be corrected if prior knowledge on the shape of the route is known. 
\begin{figure}[!t]
    \centering
    \fontfamily{cmr}\selectfont
        \includegraphics[width=\columnwidth]{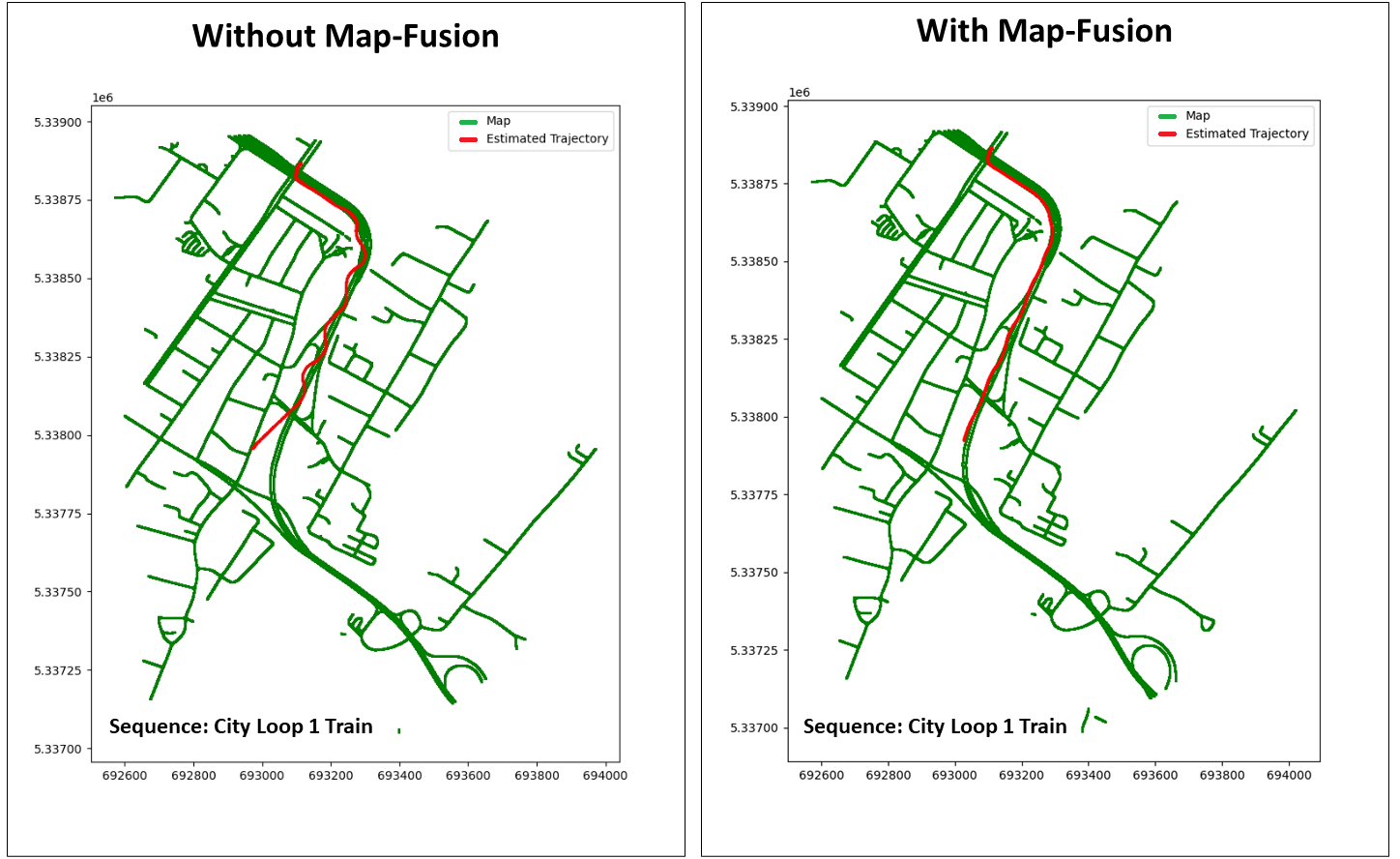}
    \caption{\textbf{Map-Fusion}: Correcting ORB-SLAM3 VIO odometry estimates for a rain + tunnel sequence in Munich from the 4Seasons Dataset. Without Map-Fusion, the odometry drifts away from lane into a different part of the city.}
    \label{representative_image}
    \vspace{-0.02\textheight}   
\end{figure}
In our proposed algorithm, the route information is used to help correct odometry estimates, whenever the estimated position of the vehicle drifts too far from the road. 
We obtain the route information from open-sourced mapping services such as OpenStreetMap (OSM) \cite{haklay_openstreetmap_2008} in the form of way-points. This makes our proposed Map-Fusion algorithm, an affordable alternative to high-definition map (HD Map). In contrast, using HD maps require either large map storage or high computation.

Thus, our proposed approach, Map-Fusion, is a sensor fusion technique that combines route information with on-board sensors to provide robust vehicle localization.
Specifically, it combines intermittent GPS readings, drifting VO or VIO estimates together with street-network based map information to perform robust localization in rain. 
Our proposed algorithm is also flexible to work with various sensor setups of the vehicle. 
Since our approach uses map information, it is susceptible to map errors and requires accurate GPS for initial localization. 
Therefore, it is to be noted that Map-Fusion should only be used in structured environments where the target vehicle mostly remains within the road region.
We utilize this structured environment to correct the drifting odometry estimate whenever it diverges from the road to prevent de-localization. 
In such scenarios, we show that our Map-Fusion algorithm helps significantly improve odometry estimates especially in rain weather and intermittent GPS segments for urban vehicular tracking. Fig. \ref{representative_image} shows Map-Fusion performing robust localization in a challenging rain and tunnel segment.

Our contributions are: (a) a novel and robust sensor fusion approach that utilizes street-network based map information to perform localization in adverse conditions, (b) an approach for processing map information that is designed for correcting localization estimates, (c) robust evaluation of our approach and baseline approaches on four geographically diverse datasets and in a real-world environment on a hardware constrained mobile robot ranging across clear and rain weather conditions. 

\section{Related Work}

\subsection{Map-based Localization}
Many existing work utilize map information to perform localization \cite{ma_exploiting_2019, guo_coarse--fine_2021, brubaker_lost_2013, shin_high_2020}. However, the map information usually comes from HD maps which could be expensive to obtain. 
Both of which could be used in conjunction with map information. 

Brubaker et al. proposed a map-based global localization approach that utilizes OSM and Visual Odometry measurements \cite{brubaker_lost_2013}. They converted the OSM road network into a graph representation and used the VO measurements to search for the most likely lane the vehicle is on in the city-wide network. Although it is able to localize with an accuracy of up to $3m$ in the KITTI dataset, it is also heavily reliant on the VO measurements. If the VO measurements starts to drift, the search for the most probable lane would be affected.
In our work, instead of inferring the lane from VO measurements, we use GPS during initialization to identify the road the vehicle starts from. Then, we use map information to correct the VO measurements if it starts to drift. 
Ma et al. \cite{ma_exploiting_2019} and Guo et al. \cite{guo_coarse--fine_2021} both uses HD map to perform localization. Their approaches involve identifying semantic objects in the scene such as road signs and traffic lights and matching with the position in the HD map. This requires highly accurate positioning of semantic objects and could be expensive to obtain. Similar to our previous work \cite{meghjani_context_2019}, which uses road contextual information to assist in performing intent and driving style prediction, we propose to use road contextual information to improve the robustness of localization estimates especially in rain scenarios.  

\subsection{Sensor Fusion Approaches}

We use GTSAM \cite{dellaert2012factor} for sensor fusion in our localization algorithm. GTSAM is a factor graph based sensor fusion approach where each sensor measurement has an associated uncertainty represented by a Gaussian distribution and the pose estimates are derived from maximizing the posterior probability of the unknown poses. The accuracy of each sensor's measurement is determined by the uncertainty given in the sensor's reading. A sensor fusion approach allows for flexibility in its usage as the user could always include additional sensors and the factor graph would still be able to solve for the optimal pose estimates. 
Another sensor fusion approach is \cite{sun_robust_2020} which proposed an Extended Kalman Filter (EKF) based sensor fusion between VO, IMU and GPS. Their approach was evaluated in GPS degraded urban areas, where GPS signals were lost for up to 4 minutes. This is similar to the problem of intermittent GPS that we explore, however, we also investigate adverse weather conditions such as rain. The difference in our approach is that we introduce map information to reduce drifts from compromised visual information caused by rain, while \cite{sun_robust_2020} innovates on the sensor fusion algorithm design.  

In contrast to the post-fusion approaches of integrating each sensor estimate, learning based multi-modal approach could perform pre-fusion on the input data level \cite{tu_ema-vio_2022, xu_rpvnet_2021, pfeuffer_optimal_2018}. This allows for one modality to complement other compromised modalities on the input level allowing learning based model to learn how to filter out compromised data. Such an approach could reduce independent error sources but is not flexible to incorporate additional sensors as retraining or restructuring of the model architecture would be required.

\subsection{Localization in Adverse Weather}

Other approaches of robust localization relies on robust sensors such as Radar and LiDAR sensors that are more resistant to adverse weather conditions \cite{hong_radarslam_2020, vizzo_kiss-icp_2023, tuna_x-icp_2024}. Hong et al. proposed a full SLAM system using only 3D Radar and showed that they were able to localize in both rain and fog \cite{hong_radarslam_2020}. Both Vizzo et al. and Tuna et al. uses LiDAR to perform robust localization \cite{vizzo_kiss-icp_2023, tuna_x-icp_2024}. Vizzo et al. focused on improving robustness across different platforms \cite{vizzo_kiss-icp_2023} while Tuna et al. improved robustness of localization in featureless environments such as underground mines and construction sites \cite{tuna_x-icp_2024}. A constraint with these approaches are that it requires either Radar or LiDAR sensors which could be expensive and bulky for tracking purposes. In our work, we focus on VO estimates for a lighter and more affordable solution to localization and tracking.

Visual Odometry and Visual Inertial Odometry are widely explored in literature, ranging from classical VO \cite{campos_orb-slam3_2021, engel_direct_2016, forster_svo_2017} to more recent learning based VO \cite{teed_droid-slam_2021, zhan_df-vo_2021, kendall_posenet_2016}. It uses a stream of image data as input to localize the pose of the camera by matching visual features between frames. Image data however is susceptible to image artefacts such as blurring and occlusions. Tan et al. evaluated various VO approaches on rainy urban datasets and found that VO alone is insufficient to localize a vehicle accurately in rain \cite{tan_evaluating_2023}. In this paper, we propose the use of map information to correct existing VO algorithms when it drifts to more accurately localize a vehicle in rainy urban scenarios.

\section{Approach} \label{approach}

\begin{figure*}[!ht]
    \centering
    \includegraphics[width=0.9\textwidth]{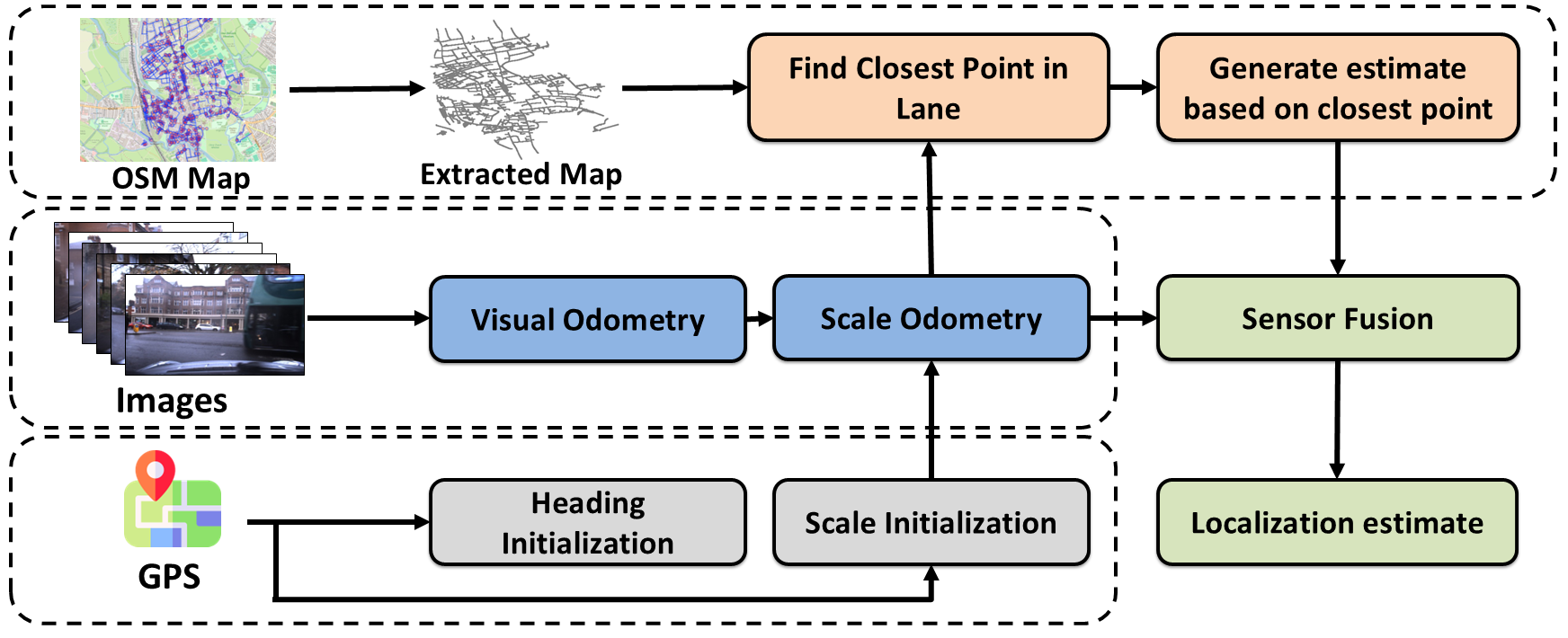} \\
    \caption{Overview of Map-Fusion.}
    \label{Map-fusion_overview}
\end{figure*}

Map-Fusion is a sensor fusion approach that uses GPS for initialization, VO or VIO for odometry estimates and map information for correcting drift in odometry. We use factor graph \cite{dellaert2012factor} for combining various sensor information and correcting the localization estimates using the map information. A factor graph was chosen as it could easily integrate the new map information and align all previous odometry estimates with the current estimate. An overview of our proposed Map-Fusion algorithm is shown in Fig. \ref{Map-fusion_overview}.

\subsection{Processing Map Information}

Map information is first obtained from OSM using a web-based data filtering tool \cite{martin_overpass_nodate}. Only lane information was extracted where each lane was treated as a $2$ way road and similar to \cite{brubaker_lost_2013}, this map was converted into a graph. Each vertex represents an intersection and each edge represents a road. Each road consists of multiple way-points given by OSM to capture the curvature of the road. Since a more fine-grained map is required for matching each odometry estimate, the way-points were linearly interpolated and smoothed using a rolling average. The connection between the segments were also refined such that the intersection between them is a smooth curve instead of a sharp edge. Lastly, each way-point was given a heading direction based on the next way-point and the last way-point follows the previous way-point's heading. The overall map is then saved to be used for localization by matching vehicle pose estimates to the map.

\subsection{Matching Vehicle Pose Estimate to Map}
At each input image frame, a visual odometry estimate can be provided from any VO algorithm. This estimate $\prescript{t-1}{}V_{t} \in SE(3)$, is the relative transform from the previous pose at time $t-1$ to the current pose at time $t$. The current pose $P_t \in SE(3)$ is parameterized as follows:
\begin{equation}
    P_t = [x,y,z,\alpha,\beta,\phi]
\end{equation}
where $x,y,z$ represents the vehicle's position and $\alpha, \beta, \phi$ represents the Euler angle representation of the vehicle's orientation. The global frame corresponds to the East, North and up direction while the relative frame corresponds to the forward, left and up direction. The closest pose on the map $R_t \in SE(3)$ to the current pose at time $t$ is defined as follows:
\begin{equation}
    \begin{split}
    \begin{gathered}
    R_t = [x,y,P_{t,z},P_{t,\alpha},P_{t,\beta},\phi] \\
    R_{t,\phi} = \arctan(\frac{R_{t,y} - R_{t-1,y}}{R_{t,x} - R_{t-1,x}})
    \end{gathered}
    \end{split}
    \label{road_heading}
\end{equation}
where $R_{t,x}$ refers to the $x$ component of $R_t$ and $P_{t,z}$ refers to the z component of $P_t$.
Given the global position of the vehicle at the previous timestep, applying the current odometry measurement provides an estimate of the vehicle's current position. 
\begin{equation}
    P_t \approx P_{t-1}\prescript{t-1}{}V_{t}
\end{equation}

In order to determine the closest point on the map to the current positional estimate, a distance metric, $D$, combining both Euclidean distance, $E$, and angular distance, $A$, was used. This metric is specified in Eq. \ref{distance_metric}. Since there are multiple points with similar angles in the map, cosine similarity was not used and the angle itself was used to better differentiate between these similar points. A higher weight was given to the angular distance such that each degree difference corresponds to $1m$ euclidean distance error. This was to ensure that the matching is based primarily on direction, such that, the turns could be better matched even if the vehicle is localized slightly off the center of the road.
\begin{equation}
    \begin{split}
    \begin{gathered}
    D = E + A \\
    A = (\arccos(2\langle q_{P_t},q_{R}\rangle^2 -1))\times\frac{180}{\pi}
    \end{gathered}
    \end{split}
    \label{distance_metric}
\end{equation}
where $q_{P_t}$ represents the quaternion representation of the orientation of pose $P_t$ and $q_{R}$ represents the quaternion representation of the orientation of a pose on the map.
The lateral distance $L$ between $P_t$ and the closest point on the map $R_t$ is calculated using Eq. \ref{lateral_distance}. $TF_t$ represents the relative transform from $R_t$ to $P_t$ and $L$ is the translation perpendicular to the direction of the road. 
\begin{equation}
    \begin{split}
    \begin{gathered}
    TF_t = R_t^{-1}P_t \\  
    L = |TF_{t,y}|
    \end{gathered}
    \end{split}
    \label{lateral_distance}
\end{equation} 

If the lateral distance exceeds the road width (approximated to be $3m$ per lane), a Gaussian prior is added to the factor graph to align the odometry estimate to the road. Since the position and orientation of the closest point is obtained from a 2D map, the roll, pitch and altitude are all set to the odometry estimate. The corresponding covariance of the Gaussian prior for the roll, pitch and altitude is set to infinity to represent uncertainty as the 2D map does not provide this information.
For the yaw parameter, the standard deviation of the Gaussian prior is empirically set as $10$ degrees to account for differences in the angle of turn at the intersections. While the Easting and Northing parameters are calculated based on the longitudinal speed $v_{lon}$ and lateral speed $v_{lat}$ of the vehicle as shown in Eq. \ref{convert_frame}. 
Since $v_{lon}$ and $v_{lat}$ are in the relative frame of the vehicle, a conversion is required to get the global frame representation of the covariance.
Using Eigen-decomposition, the covariance in global coordinates can be decomposed into its eigenvectors corresponding to the longitudinal and lateral direction. Thus, given the longitudinal and lateral eigenvector $e_1$, $e_2$ and eigenvalues $\lambda_1$, $\lambda_2$, the global coordinate covariance $\Sigma$ can be determined.
\begin{displaymath}
    \begin{split}
    \begin{gathered}
    e_1 = \begin{bmatrix}P_{t,x} - P_{t-1,x} & P_{t,y} - P_{t-1,y}\end{bmatrix} \\
    e_2 = \begin{bmatrix}P_{t,y} - P_{t-1,y} & P_{t-1,x} - P_{t,x} \end{bmatrix} \\
    \lambda_1 = |v_{lon}| \\
    \lambda_2 = |v_{lat}|
    \end{gathered}
    \end{split}
\end{displaymath}
\begin{equation}
    \begin{split}
    \Sigma = 
    \begin{bmatrix}
    \vert & \vert \\
    e_1   & e_2   \\
    \vert & \vert
    \end{bmatrix}	
    \begin{bmatrix}
    \lambda_1 & 0 \\
    0   & \lambda_2
    \end{bmatrix}
    \begin{bmatrix}
    \vert & \vert \\
    e_1   & e_2   \\
    \vert & \vert
    \end{bmatrix}^{-1}
    \end{split}
    \label{convert_frame}
\end{equation}

\subsection{GPS Initialization and Scaling}
In order to align the odometry from local coordinate frame to the global frame, a series of GPS coordinates are required. This GPS sequence was used to estimate both vehicle heading and scale of the localization estimates. The heading is estimated by taking the arc-tangent of consecutive GPS coordinates similar to estimating heading of map poses in Eq. \ref{road_heading}. 
In the process of determining the scale using GPS, three thresholds were used. One threshold sets a minimum distance travelled by the vehicle before estimating scale which is represented as $md$ and another threshold sets the minimum average speed for the vehicle in $km/h$, denoted as $ms$. These two thresholds are used together in Eq. \ref{gps_vo_count} to determine the number of GPS measurements $C_{gps}$ and number of VO measurements $C_{vo}$ required to travel the threshold distance at the threshold speed. $f_{vo}$ and $f_{gps}$ refers to the frequency of VO and GPS measurements.
\begin{equation}
    \begin{split}
    \begin{gathered}
    C_{gps} = \lceil \frac{md \times 3.6 \times f_{gps}}{ms} \rceil \\
    C_{vo} = round(\frac{C_{gps} \times f_{vo}}{f_{gps}})
    \end{gathered}
    \end{split}
    \label{gps_vo_count}
\end{equation}
At every timestep, the distance travelled by GPS measurements and the distance travelled by VO measurements are accumulated. When the GPS distance exceeds the minimum distance threshold and the number of GPS and VO measurements do not exceed $C_{gps}$ and $C_{vo}$ respectively, then the scale $S$ is estimated using Eq. \ref{scale}.
\begin{equation}
    \begin{split}
    \begin{gathered}
    S = \frac{\norm{d_{gps}}^2}{\norm{d_{vo}}^2} \times \frac{t_{vo}}{t_{gps}}
    \end{gathered}
    \end{split}
    \label{scale}
\end{equation}
where $d_{gps}$ and $d_{vo}$ represents the distance accumulated by GPS and VO measurements. $t_{gps}$ and $t_{vo}$ represents the duration in seconds to accumulate $d_{gps}$ and $d_{vo}$ respectively.
The last threshold sets the number of times scale estimation is repeated to average out the errors. Median was used instead of mean to filter outliers as we found that it was easy for the scale estimation to deviate greatly with a few highly uncertain GPS measurements. A higher number of repeated scale estimation could result in a more reliable scale at the cost of a longer initialization time. In our experiments, we found that $50$ samples were sufficient to determine the scale reliably. 

\section{Experiments}
Our proposed Map-Fusion algorithm was extensively tested on $4$ datasets across both in clear and rain weather scenarios and across multiple countries. We experimented on the datasets with and without Map-Fusion for both a VO approach and a VIO approach and showed that on average it improved both the VO and VIO results.
The parameters are tuned empirically and the exact values alongside the implementation of the Map-Fusion algorithm is open-sourced\footnote{https://gitlab.com/marvl/map-fusion}. 

\subsection{Datasets and Evaluation}
\begin{figure}[ht]
    \centering
    \newcolumntype{C}{>{\centering\arraybackslash}X}
    \begin{tabularx}{\columnwidth}{CC}
    \fontfamily{cmr}\selectfont
        \includegraphics[width=0.42\columnwidth]{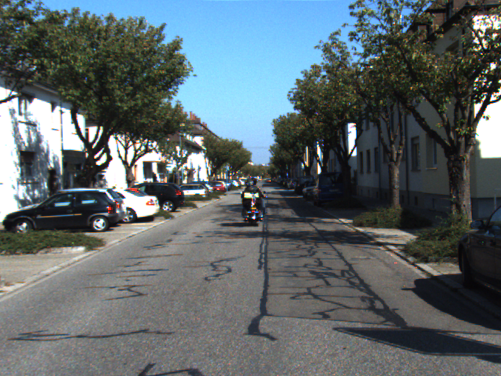} &
        \includegraphics[width=0.42\columnwidth]{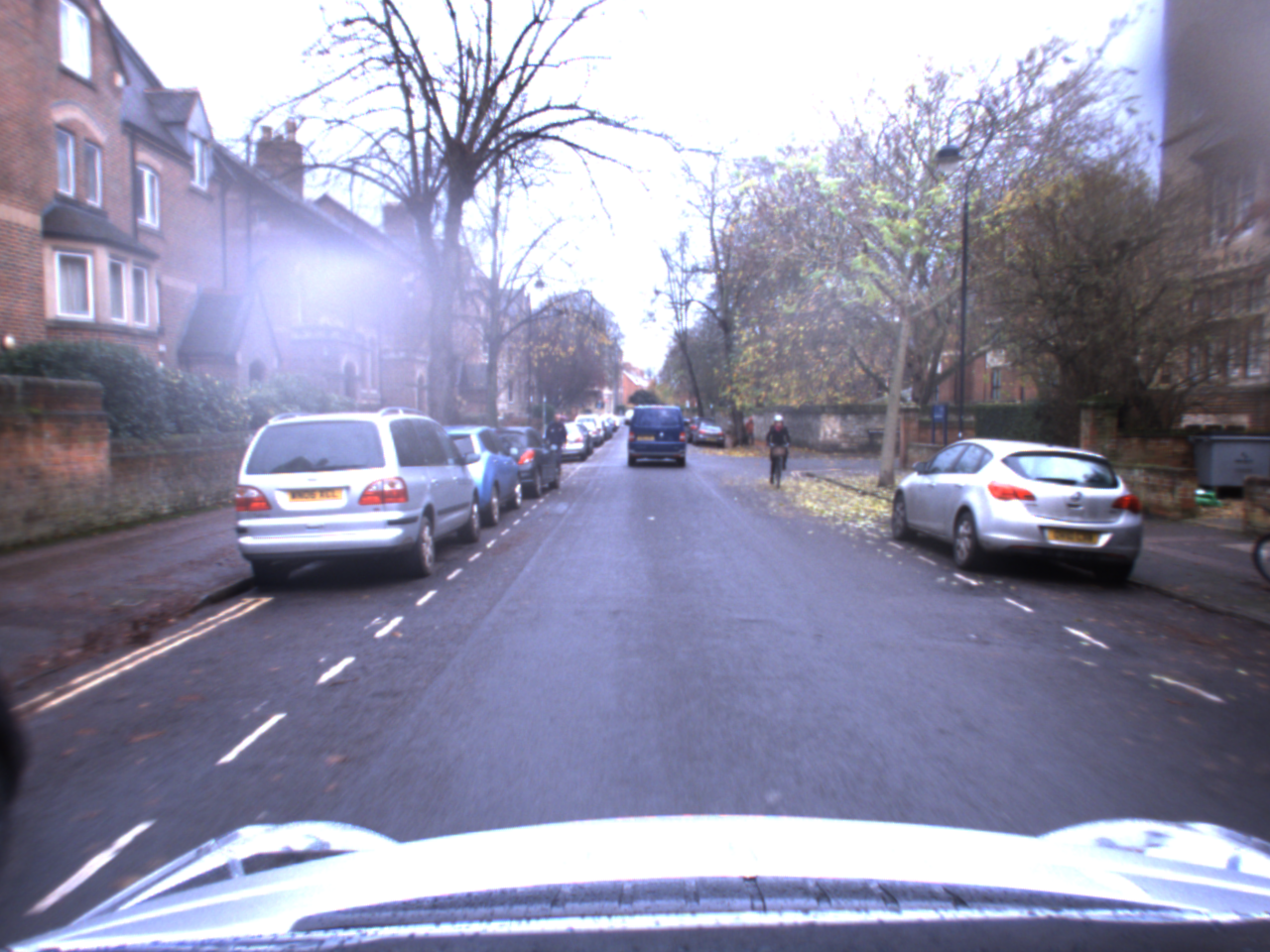}\\
        \scriptsize{KITTI} & \scriptsize{Oxford Robotcar} \\
        \includegraphics[height=.3\columnwidth,width=0.42\columnwidth]{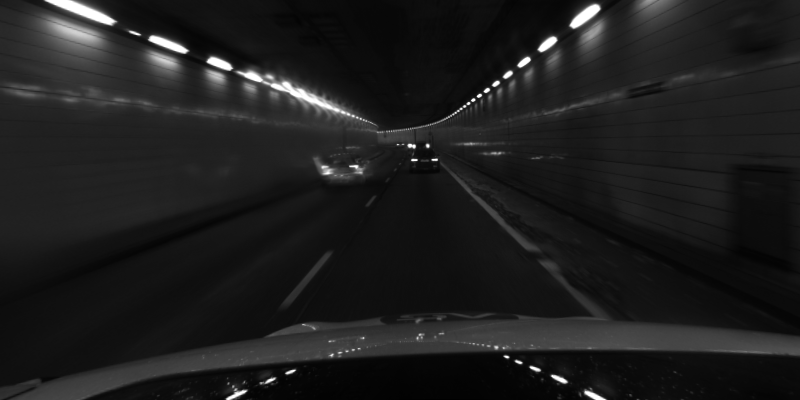} &
        \includegraphics[height=.3\columnwidth,width=0.42\columnwidth]{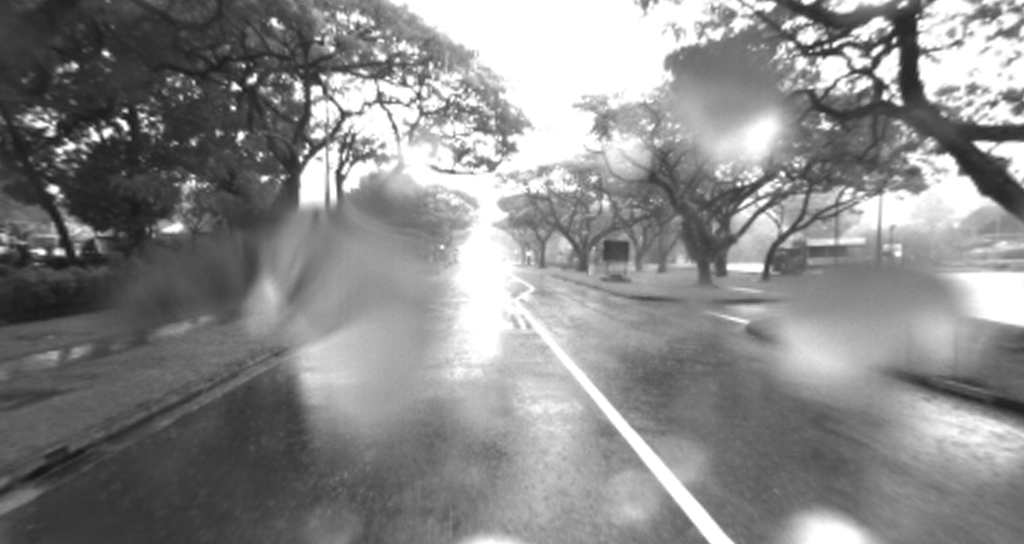}  \\
        \scriptsize{4Seasons} & \scriptsize{Singapore} \\
    \end{tabularx}
    \caption{Sample images from the four datasets.}
    \label{fig:datasets}
\end{figure}

The $4$ datasets used are: (a) KITTI Dataset (for clear weather) \cite{geiger_are_2012}, (b) Oxford Robotcar Dataset (for rain weather) \cite{maddern_1_2017, maddern_real-time_2020}, (c) 4Seasons Dataset (for rain + tunnel) \cite{wenzel_4seasons_2020} and (d) an internal dataset collected in Singapore (for heavy rain). Fig. \ref{fig:datasets} shows sample images from each of the datasets. Only the rain sequences and $1$ clear sequence for each route from Oxford Robotcar and 4Seasons Datasets are used. The sequences from Oxford Robotcar are also cut short to match the shortest sequence such that the errors could be fairly compared against each other. In the Oxford Robotcar Dataset, VIO was not evaluated as there are no raw IMU readings provided. For the KITTI Dataset, there were multiple time gaps in the raw IMU readings. As such, the IMU reading was extrapolated using a rolling average of the past $50$ readings for such gaps. In the 4Seasons dataset, the tunnel sequence was cut short to right after the vehicle passes the tunnel to analyze the localization error within the tunnel.

For all $4$ datasets, the output poses from Map-Fusion (in the global frame) are aligned (6DoF) to the ground truth before evaluation. In the following evaluation, all output poses and ground truth poses are projected onto the 2D plane before calculating the Absolute Trajectory Error (ATE) \cite{sturm_benchmark_2012}.

\subsection{Experimental Setup}
We evaluated our proposed Map-Fusion algorithm using a VO and a VIO approach. We used DROID-SLAM \cite{teed_droid-slam_2021} for the VO approach and ORB-SLAM3 \cite{campos_orb-slam3_2021} for the VIO approach. Both DROID-SLAM and ORB-SLAM3 were chosen as they provide a stereo camera mode and have good scale consistency throughout the route. This is also to evaluate Map-Fusion on different types of odometry approaches where DROID-SLAM is a learning based approach while ORB-SLAM3 is a classical approach. In both approaches, the stereo option was used. 
For DROID-SLAM, the GPU memory usage was restricted to $12GB$ to prevent out of memory errors for long sequences. Also, post-processed global optimization was disabled to ensure localization results were obtained in real-time. 
For ORB-SLAM3, since the results varies significantly each run, the evaluation was carried out $3$ times for each sequence and the average is reported. If it de-localizes any $1$ out of the $3$ runs, it is reported as de-localized. For all approaches, de-localization is defined when ATE errors exceed $20m$. This threshold was chosen as it was the widest road observed in all the datasets evaluated. Specifically for ORB-SLAM3, a de-localization is also defined as when ORB-SLAM3 resets its local map. 
The VO and VIO algorithms were executed independently and were used as baseline algorithms for comparison with our proposed Map-Fusion algorithm. 
The VO and VIO algorithms without Map-Fusion were initialized with GPS but they do not use map information. This is to evaluate the effects of the map information separately from the GPS scaling and pose initialization. 
For the KITTI and Oxford Robotcar datasets, we use GPS only for initialization. Using the initialized segment, our proposed approach is able to robustly keep track of the vehicle in light rain scenarios. However, in challenging scenarios such as the tunnel segment in the 4Seasons dataset and heavy rain in the Singapore dataset, the visual input is severely compromised. As such, we introduce intermittent GPS measurements (every $5$ seconds) to reduce drift.

We used GTSAM \cite{dellaert2012factor} as the sensor fusion approach. GTSAM is a factor graph based sensor fusion approach which assumes that every measurement taken is Gaussian with its own covariance matrix to define the uncertainty of the measurement. Since the VO approaches do not come with a predicted covariance alongside its estimated pose, two covariance matrices were set for the odometry measurements. One covariance matrix was set to be highly uncertain before initialization when scaling is not complete and the other with higher certainty after initialization when scaling is completed. Similarly for the GPS measurement, the covariance matrix was set to a predefined value as the dataset does not provide covariance values. Although the odometry approaches provide camera frame absolute pose, the relative odometry pose was used in GTSAM to model the propagation of uncertainty. Such an approach allows for the correction of all previous odometry poses to fit a new map-based alignment. However, it could also cause uncertainty to propagate without bound.
To counter this, the odometry estimates are given as priors to GTSAM to reduce uncertainty whenever the standard deviation of the positional estimates (modelled as Gaussian Distributions) exceeds the road width. This controls the propagation of uncertainty with each odometry estimate. 

For each of the dataset, a single map was obtained to cover every sequence evaluated. The largest map extracted covers an approximate area of $10$km by $15$km used for the KITTI dataset. Only the roads were extracted along with its meta-information such as number of lanes and whether it is "one\_way". All the lanes were converted into two way streets to reduce map errors and to allow for more possible movements such as moving in reverse. Only one particular lane in the map for the Oxford Robotcar Dataset was manually adjusted to have its number of lanes set to $6$. This is to correct a map error shown in Fig. \ref{potential errors in map} where the map is missing lane information and is unable to properly represent the size of the parking space. Since the width of the parking space was approximately $20m$, using an approximation of $3m$ per lane, $6$ is the closest number of lanes to $20m$ without exceeding it. For all other roads with missing lane information, the number of lanes are set to $1$. All searches on the map runs in real-time for online tracking of the vehicle despite the large size of the map. This is possible because the initial position is obtained during initialization which allows the subsequent search space to be narrowed down to be within a $20m$ radius of the initial match. 

\begin{table*}[ht]
\caption{Absolute Trajectory Errors(ATE) across KITTI sequences. + Map-Fusion refers to the integration of the map information. All approaches were initialized with GPS. No lc refers to no loop closure. Reduction in errors $>3m$ are bolded. x represents de-localization.}
\centering
\begin{tabularx}{\linewidth}{ | c | >{\centering\arraybackslash}X > {\centering\arraybackslash}X >{\centering\arraybackslash}X > {\centering\arraybackslash}X >{\centering\arraybackslash}X >{\centering\arraybackslash}X >{\centering\arraybackslash}X >{\centering\arraybackslash}X >{\centering\arraybackslash}X >
{\centering\arraybackslash}X| >
{\centering\arraybackslash}X|}
 \hline
 ATE(m) & \multicolumn{11}{|c|}{KITTI Dataset}\\
 \hline
 
 Approaches & 00 & 01 & 02 & 04 & 05 & 06 & 07 & 08 & 09 & 10 & Avg \\
 \hline
 DROID-SLAM & 2.98 & x & 5.77 & 0.65 & 2.29 & 2.10 & 1.01 & 4.77 & 3.44 & 1.59 & 2.73\\
 DROID-SLAM + Map-Fusion & 2.20 & x & 3.18 & 0.65 & 3.14 & 2.60 & 1.02 & 2.60 & 2.61 & 1.12 & 2.12\\
 \hline
 Reduction in Error & 0.78 & NA & 2.59 & 0.00 & -0.85 & -0.50 & -0.01 & 2.17 & 0.83 & 0.47 & 0.61\\
 \hline
 ORB-SLAM3 VIO & x & 3.93 & 13.43 & 0.90 & 2.60 & x & 2.32 & 10.37 & 6.22 & 2.67 & 5.30\\
 ORB-SLAM3 VIO + Map-Fusion & x & 2.88 & 4.61 & 0.90 & 2.36 & x & 1.19 & 3.60 & 3.52 & 3.18 & 2.78\\
 \hline
 Reduction in Error & NA & 1.05 & \textbf{8.82} & 0.00 & 0.24 & NA & 1.13 & \textbf{6.77} & 2.70 & -0.51 & 2.52\\
 \hline
 ORB-SLAM3 VIO (no lc)& x & 3.68 & x & 0.88 & 7.89 & x & 2.32 & 10.40 & 7.94 & 2.65 & 5.11\\
 ORB-SLAM3 VIO (no lc) + Map-Fusion & \textbf{8.35} & 2.93 & x & 0.88 & 2.31 & x & 1.28 & 3.52 & 3.08 & 3.07 & 2.44\\
 \hline
 Reduction in Error & NA & 0.75 & NA & 0.00 & \textbf{5.58} & NA & 1.04 & \textbf{6.88} & \textbf{4.86} & -0.42 & 2.67\\

 \hline
\end{tabularx}
\label{table:kitti}
\end{table*}

\begin{table*}[ht]
\caption{Absolute Trajectory Errors(ATE) across Oxford Robotcar sequences. + Map-Fusion refers to the integration of the map information. Reduction in errors $>3m$ are bolded. All approaches were initialized with GPS. x represents de-localization.}
\centering
\begin{tabularx}{\linewidth}{ | c | >{\centering\arraybackslash}X > {\centering\arraybackslash}X >
{\centering\arraybackslash}X >{\centering\arraybackslash}X| >{\centering\arraybackslash}X|}
 \hline
 ATE(m) & \multicolumn{5}{|c|}{Oxford Robotcar Dataset} \\
 \hline
 
 Approaches & 2014-12-09-13-21-02 (Clear) & 2015-10-29-12-18-17 (Rain) & 2014-11-25-09-18-32 (Rain) & 2014-11-21-16-07-03 (Rain + Night) & Average \\
 \hline
 DROID-SLAM & 10.19 & 10.36 & 18.18 & x & 12.91\\
 DROID-SLAM + Map-Fusion & 3.44 & 4.02 & 10.07 & x & 5.84\\
 \hline
 Reduction in Error & \textbf{6.75} & \textbf{6.34} & \textbf{8.11} & NA & \textbf{7.07}\\

 \hline
\end{tabularx}
\label{table:oxford}
\end{table*}

\begin{table*}[ht]
\caption{Absolute Trajectory Errors(ATE) across  4Seasons sequences. + Map-Fusion refers to the integration of the map information. All approaches were initialized with GPS. No lc refers to no loop closure. x represents de-localization.}
\centering
\begin{tabularx}{\linewidth}{ |c | >{\centering\arraybackslash}X >{\centering\arraybackslash}X >{\centering\arraybackslash}X  >{\centering\arraybackslash}X| >
{\centering\arraybackslash}X|}
 \hline
 ATE(m) & \multicolumn{5}{|c|}{4Seasons} \\
 \hline
 Approaches & neighborhood\_2 \_train (Clear) & neighborhood\_3 \_train (Rain) & city\_loop\_3 \_train (Clear + tunnel) & city\_loop\_1 \_train (Rain + tunnel) & Average \\
 \hline
 DROID-SLAM + GPS every 5s & 0.50 & 0.56 & x & x & 0.53\\
 DROID-SLAM + Map-Fusion + GPS every 5s & 0.49 & 0.57 & x & x & 0.53\\
 \hline
 Reduction in Error & 0.01 & -0.01 & NA & NA & 0.00\\
 \hline
 ORB-SLAM3 VIO + GPS every 5s & 0.51 & 0.64 & 5.49 & x & 2.21\\
 ORB-SLAM3 VIO + Map-Fusion + GPS every 5s & 0.50 & 0.60 & 3.80 & x & 1.63\\
 \hline
 Reduction in Error & 0.01 & 0.04 & 1.69 & NA & 0.58\\
 \hline
 ORB-SLAM3 VIO (no lc) + GPS every 5s & 0.52 & 0.63 & x & x & 0.58\\
 ORB-SLAM3 VIO (no lc) + Map-Fusion + GPS every 5s & 0.51 & 0.58 & x & x & 0.55\\
 \hline
 Reduction in Error & 0.01 & 0.05 & NA & NA & 0.03\\

 \hline
\end{tabularx}
\label{table:4seasons}
\end{table*}

\begin{table*}[ht]
\caption{Absolute Trajectory Errors(ATE) across Singapore sequences. + Map-Fusion refers to the integration of Map information. All approaches were initialized with GPS. x represents de-localization.}
\centering
\begin{tabularx}{\linewidth}{ | c | >{\centering\arraybackslash}X > {\centering\arraybackslash}X >
{\centering\arraybackslash}X |>{\centering\arraybackslash}X|}
 \hline
 ATE(m) & \multicolumn{4}{|c|}{Singapore Dataset} \\
 \hline
 
 Approaches & Zero (Rain) & One (Rain) & Five (Rain) & Average \\
 \hline
 DROID-SLAM + GPS every 5s & 7.88 & 2.82 & 1.57 & 4.09\\
 DROID-SLAM + Map-Fusion + GPS every 5s & 7.02 & 2.58 & 1.19 & 3.60\\
 \hline
 Reduction in Error & 0.86 & 0.24 & 0.38 & 0.49\\

 \hline
\end{tabularx}
\label{table:singapore}
\end{table*}
\subsection{Quantitative Results}
Table \ref{table:kitti} reports the ATE for the KITTI Dataset recorded in clear weather. 
Localization estimates with larger than $20m$ error is considered to be delocalized and are marked with an x. 
The delocalized sequence is thus not included in the average. 
We observe that ORB-SLAM3 VIO performed worse compared to DROID-SLAM VO. This is likely due to missing IMU information for certain sequences.
Comparing with and without Map-Fusion for DROID-SLAM VO and ORB-SLAM3 VIO without loop closure, Map-Fusion on average improves the ATE by $0.61$m and $2.67m$, respectively. Also, Map-Fusion improves the ATE for ORB-SLAM3 VIO with loop closure by $2.53m$. 
Specifically, our proposed Map-Fusion algorithm, was useful in recovering de-localized sequences caused by lateral drifts.
An example is sequence $00$ where ORB-SLAM3 VIO (no loop closure) was de-localized due to lateral drifts and we successfully corrected it using the map information. 
Our approach however, is limited by the longitudinal drift correction capabilities of the underlying algorithm. 
For example in sequence $01$, which has a lack of temporal visual features to determine longitudinal motion of the vehicle, a pure VO algorithm such as DROID-SLAM will fail. 
In contrast, a VIO algorithm such as ORB-SLAM3 is able to track the longitudinal motion of the vehicle.
Thus, our proposed Map-Fusion algorithm was delocalized for DROID-SLAM and improved ATE for ORB-SLAM3 VIO.
The sequences with negative reduction in errors (DROID-SLAM sequence 05, 06, 07 and ORB-SLAM3 VIO sequence 10) could be attributed to poor matching to the map causing the correction to be inaccurate when the odometry results without Map-Fusion were already accurate ($<3m$ error). Since Map-Fusion was able to improve both ORB-SLAM3 VIO with and without loop closure, it shows that it is also generalizable to odometry approaches that perform their own loop closure. 

Table \ref{table:oxford} shows the ATE for the Oxford Robotcar Dataset which evaluates the performance of localization in rain weather. 
Map-Fusion shows promising results, reducing errors for $3$ out of the $4$ sequences. 
On average, Map-Fusion reduced errors by $7.07m$ which shows that it is effective in rain conditions and helps improve robustness of localization. 
For sequence $2014$-$11$-$21$-$16$-$07$-$03$, due to rain and night conditions, the accumulated drift was too large for Map-Fusion to correct thus both DROID-SLAM and Map-Fusion delocalized.

Table \ref{table:4seasons} shows the ATE for the 4Seasons Dataset. 
Given the challenging sequences in this dataset, GPS was provided every $5$s to reduce drift. It should be noted that there are no GPS measurements within the tunnel. 
DROID-SLAM de-localizes upon entering the tunnel while ORB-SLAM3 with additional IMU measurements was able to continue localizing throughout the tunnel. Map-Fusion corrected odometry drift within the tunnel for sequence city\_loop\_3\_train, reducing errors by $1.69m$. It also helped correct drift in the sequence city\_loop\_1\_train but due to a de-localization in one of the $3$ runs, it is recorded as delocalized. This is further discussed in the qualitative evaluation section.

The Singapore dataset was recorded in heavy rain, making visual odometry particularly difficult. Thus, GPS was also provided every 5s and the results are shown in Table \ref{table:singapore}. There is a reduction in error for every sequence, averaging $0.49$m across all sequences.

Across all datasets and odometry approaches, averaging across sequences, Map-Fusion reduces localization errors and showcasing its generalizability for different odometry approaches in challenging urban driving scenarios.

\subsection{Qualitative Results}
\begin{figure}[h]
    \centering
    \fontfamily{cmr}\selectfont
        \includegraphics[width=.48\columnwidth]{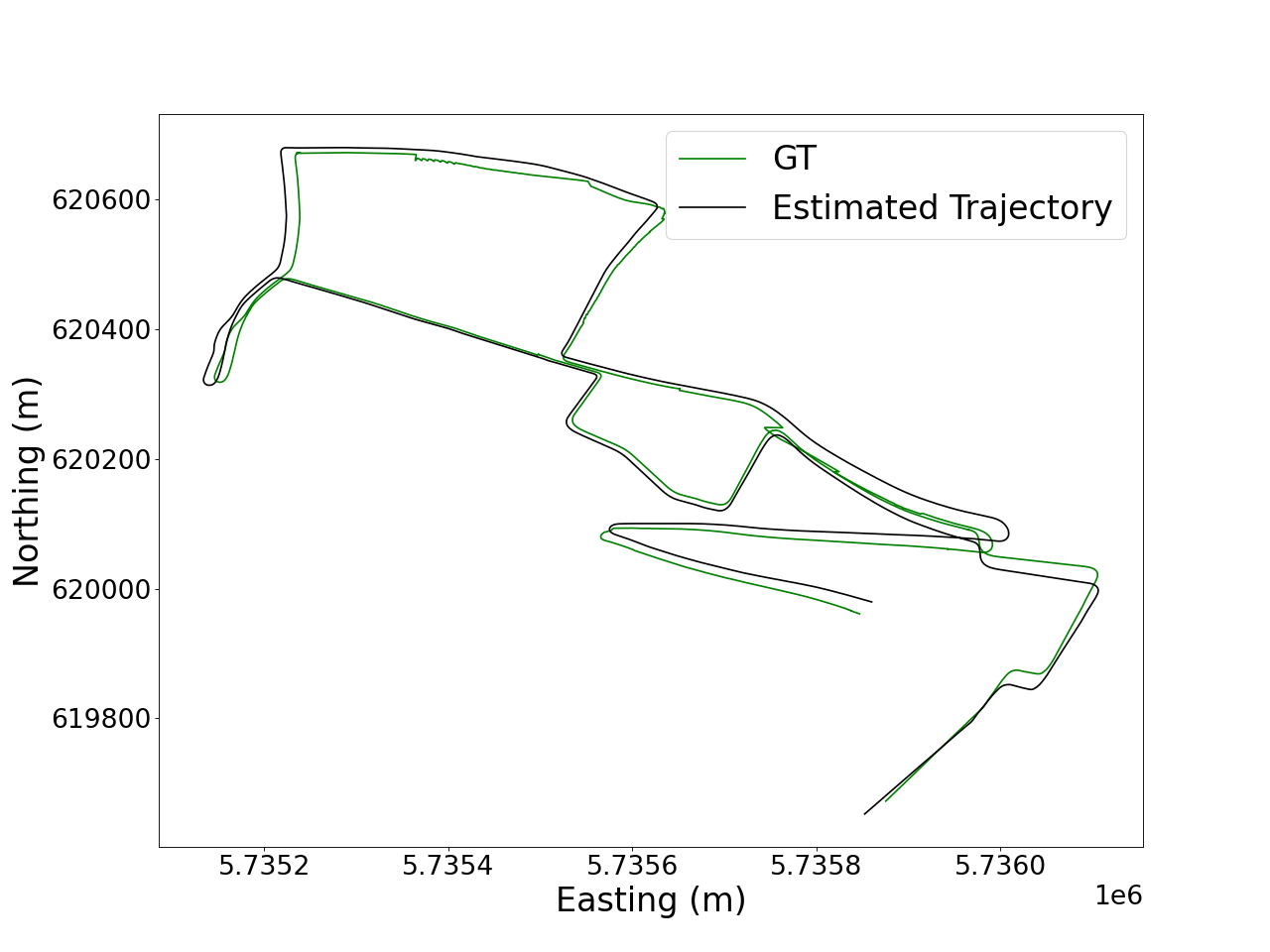}
        \includegraphics[width=.48\columnwidth]{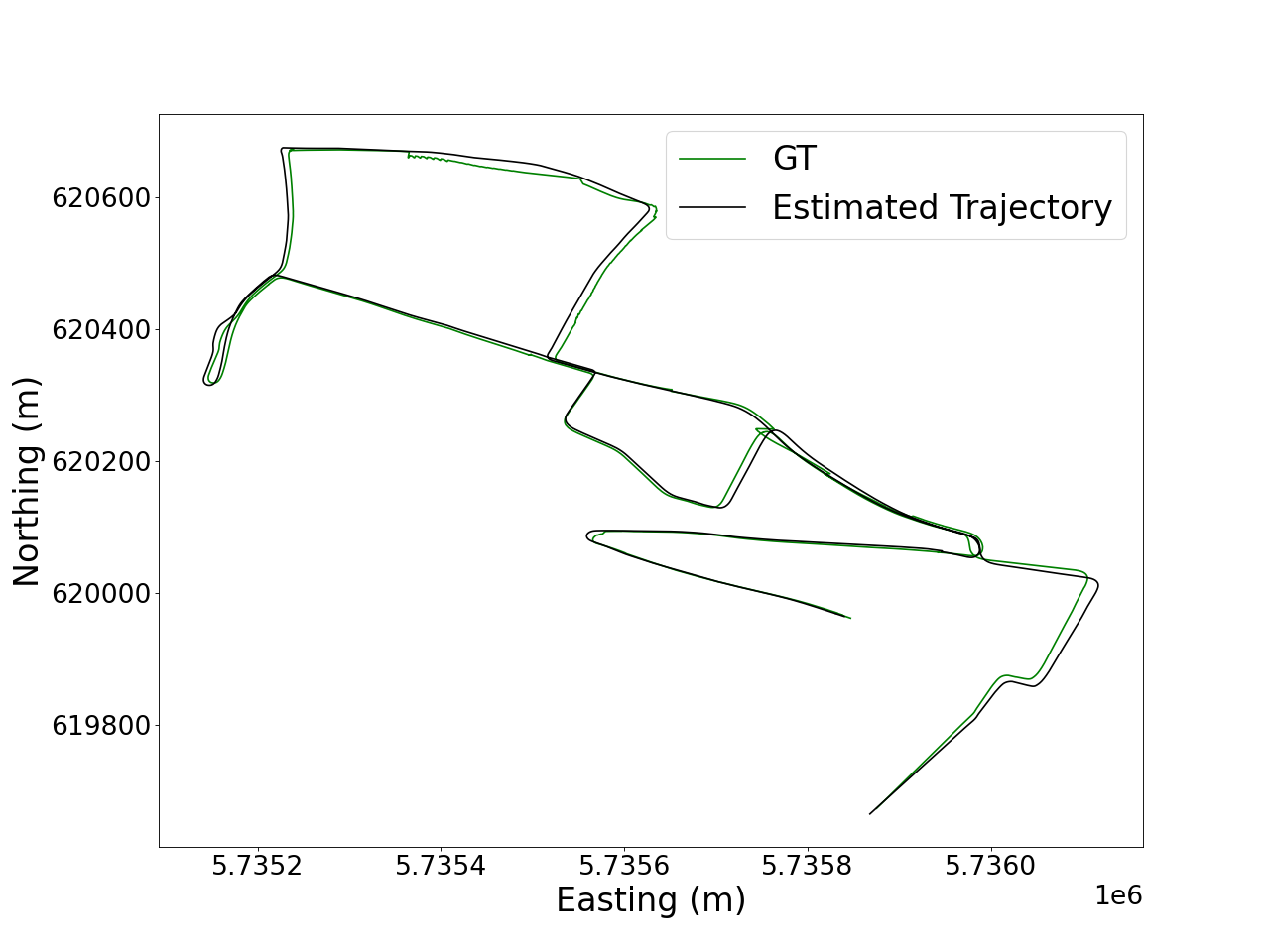} \\
        \includegraphics[width=.48\columnwidth]{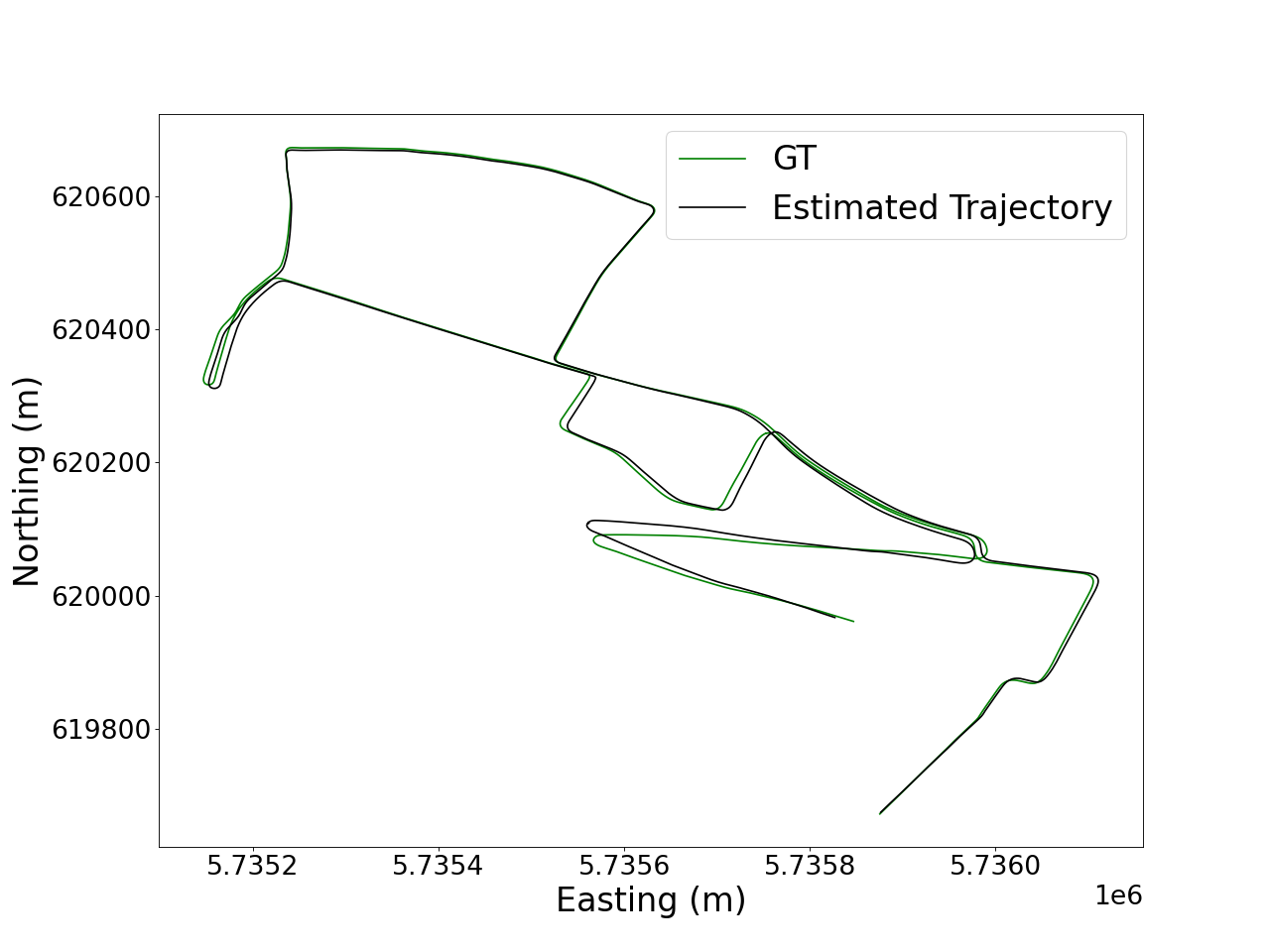}
        \includegraphics[width=.48\columnwidth]{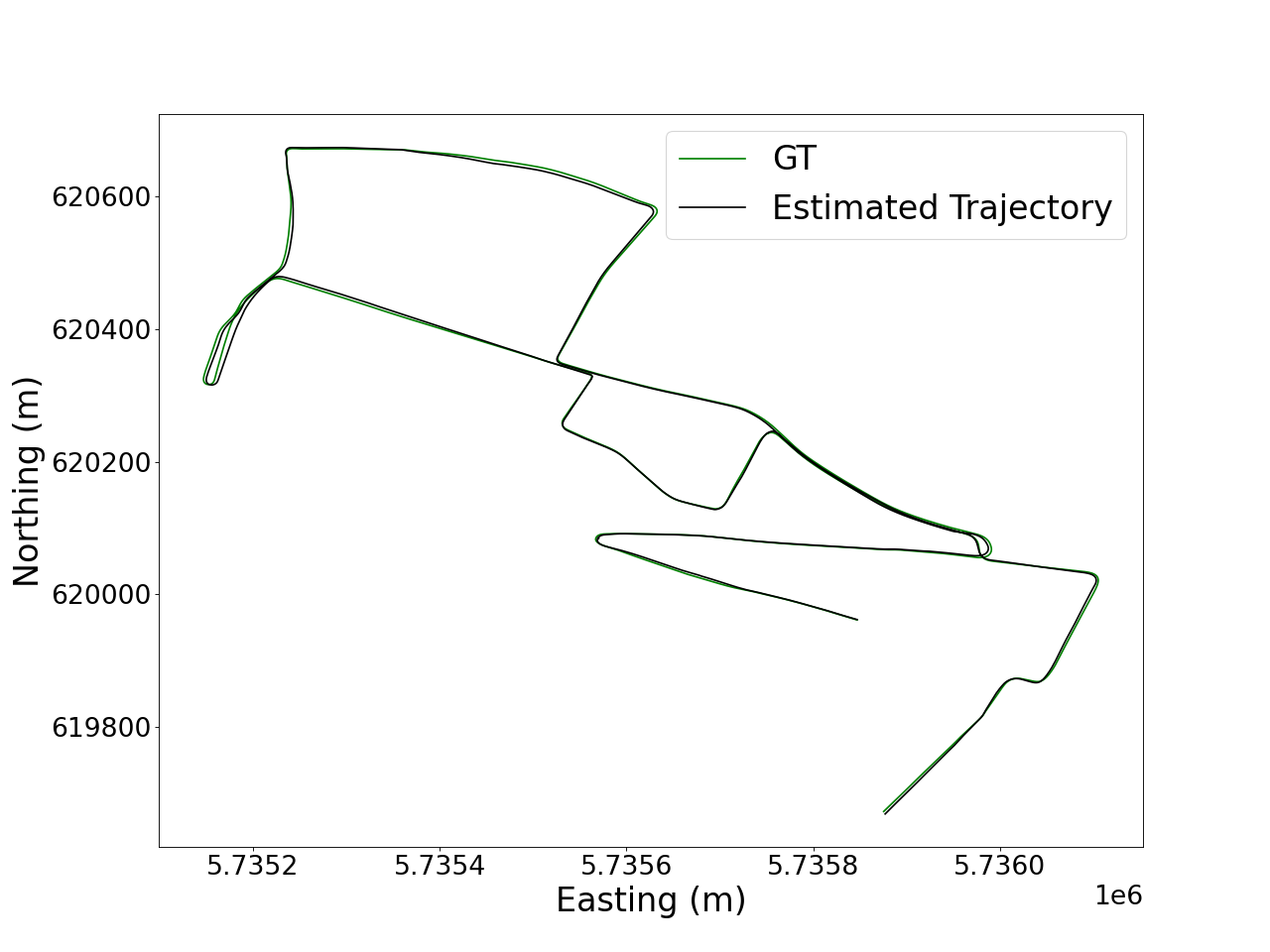} \\
        \includegraphics[width=.48\columnwidth]{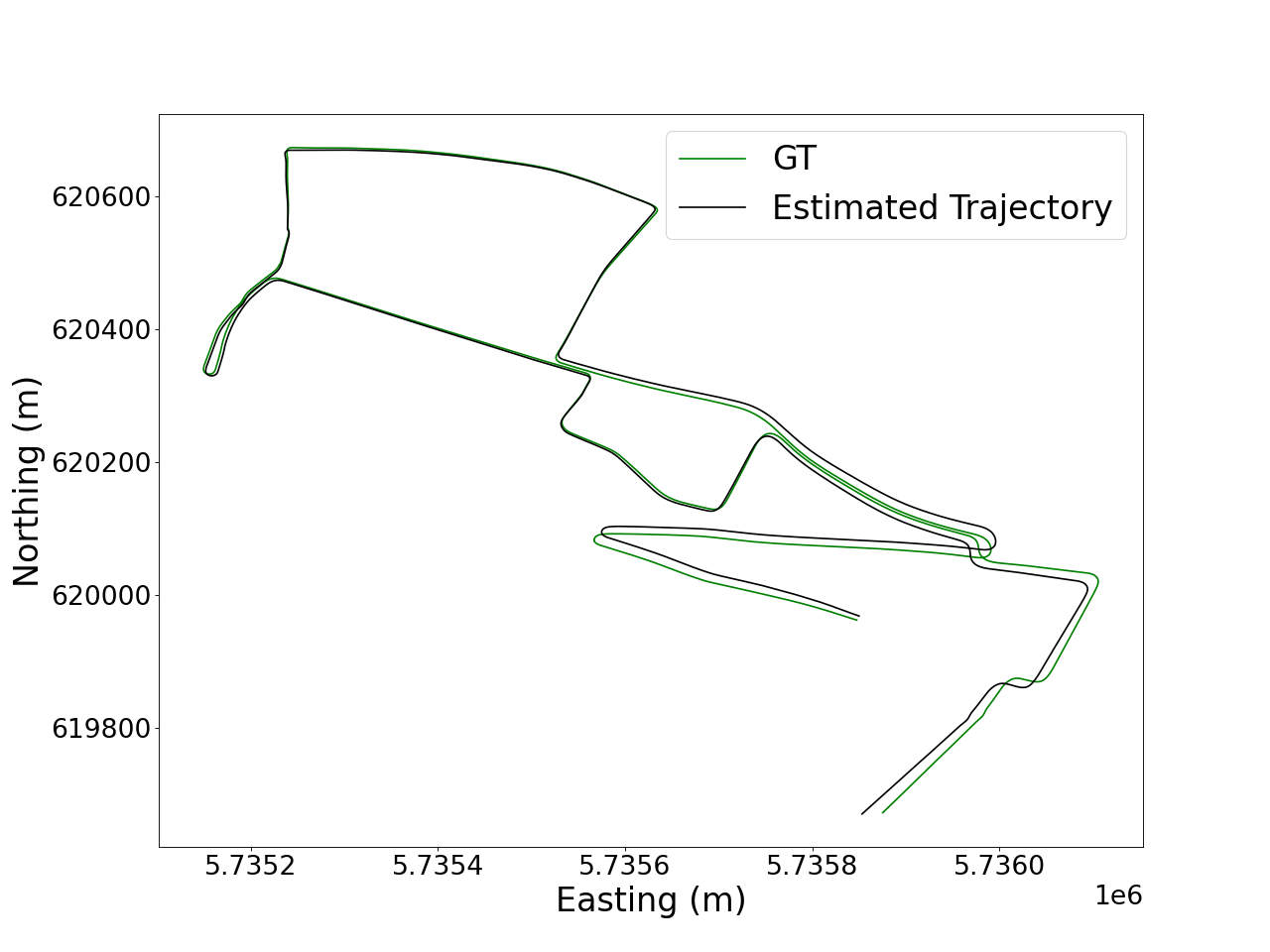}
        \includegraphics[width=.48\columnwidth]{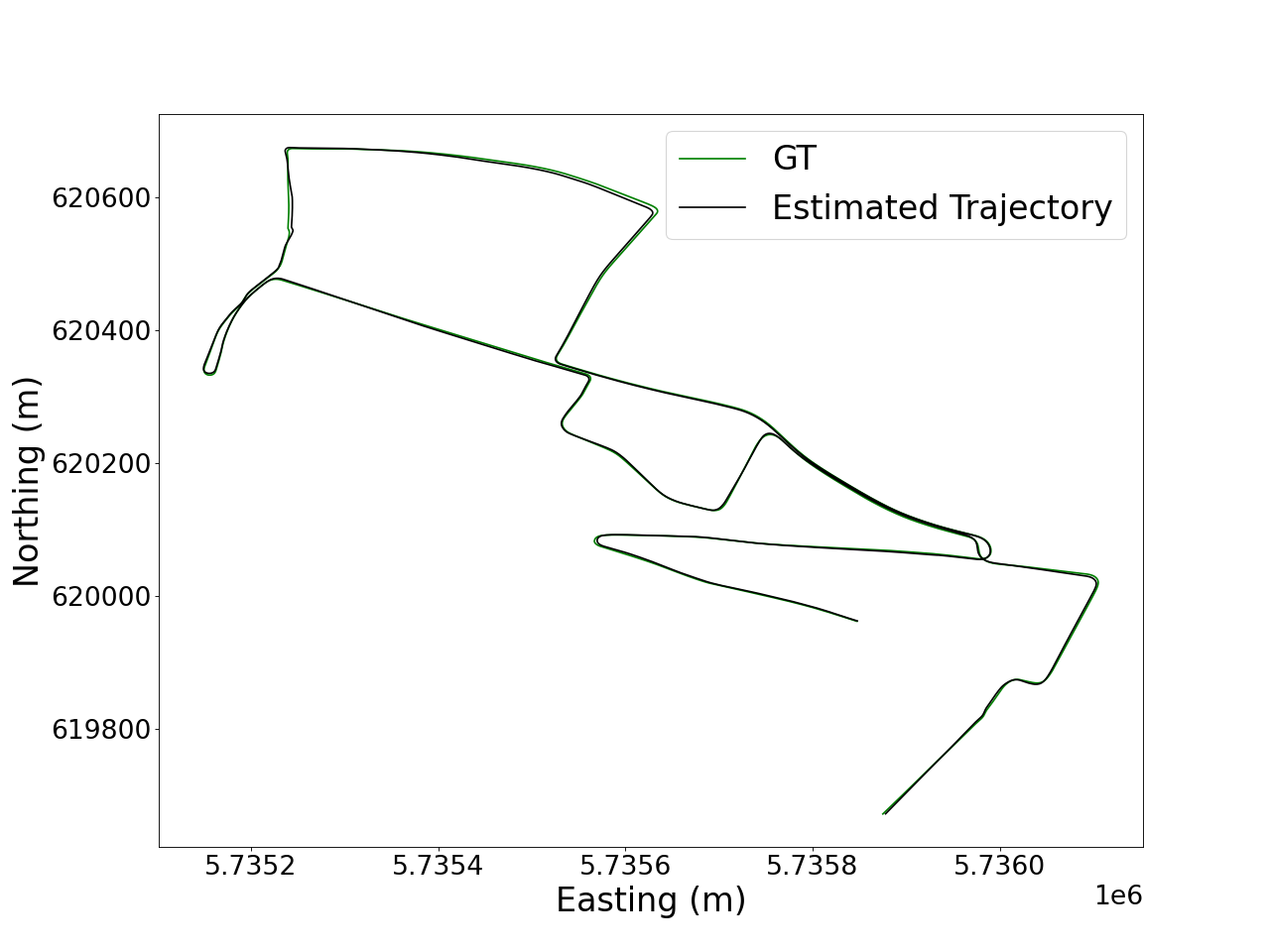} \\
        \scriptsize{DROID-SLAM} \hspace{0.12\textwidth} \scriptsize{DROID-SLAM}\\
        \scriptsize{\hphantom{.}} \hspace{0.22\textwidth}
        \scriptsize{+ Map-Fusion} \\
    \caption{Output trajectories of DROID-SLAM (left) and DROID-SLAM + Map-Fusion(right) for Oxford Robotcar sequences 2014-11-25-09-18-32 (top), 2015-10-29-12-18-17 (middle),  2014-12-09-13-21-02 (bottom)}
    \label{qualitative_correction}
\end{figure}

\begin{figure}[h]
    \centering
    \fontfamily{cmr}\selectfont
        \includegraphics[width=.48\columnwidth]{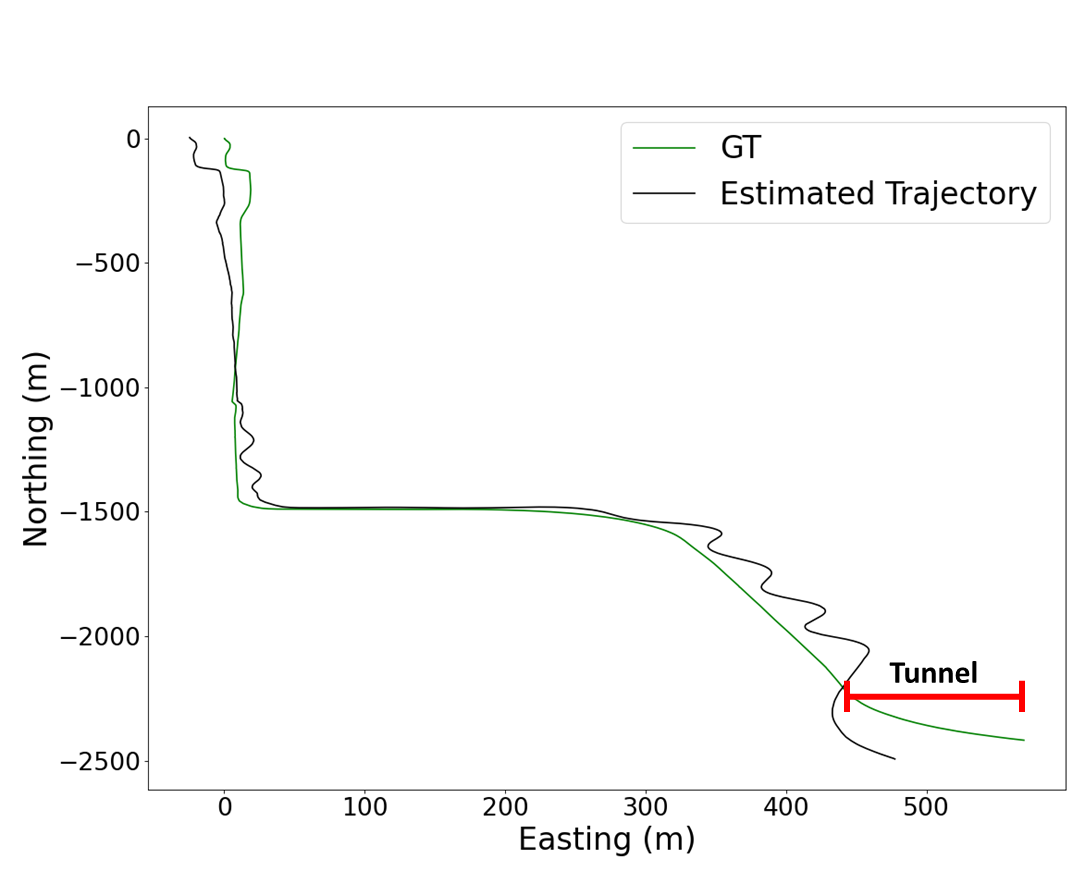}
        \includegraphics[width=.48\columnwidth]{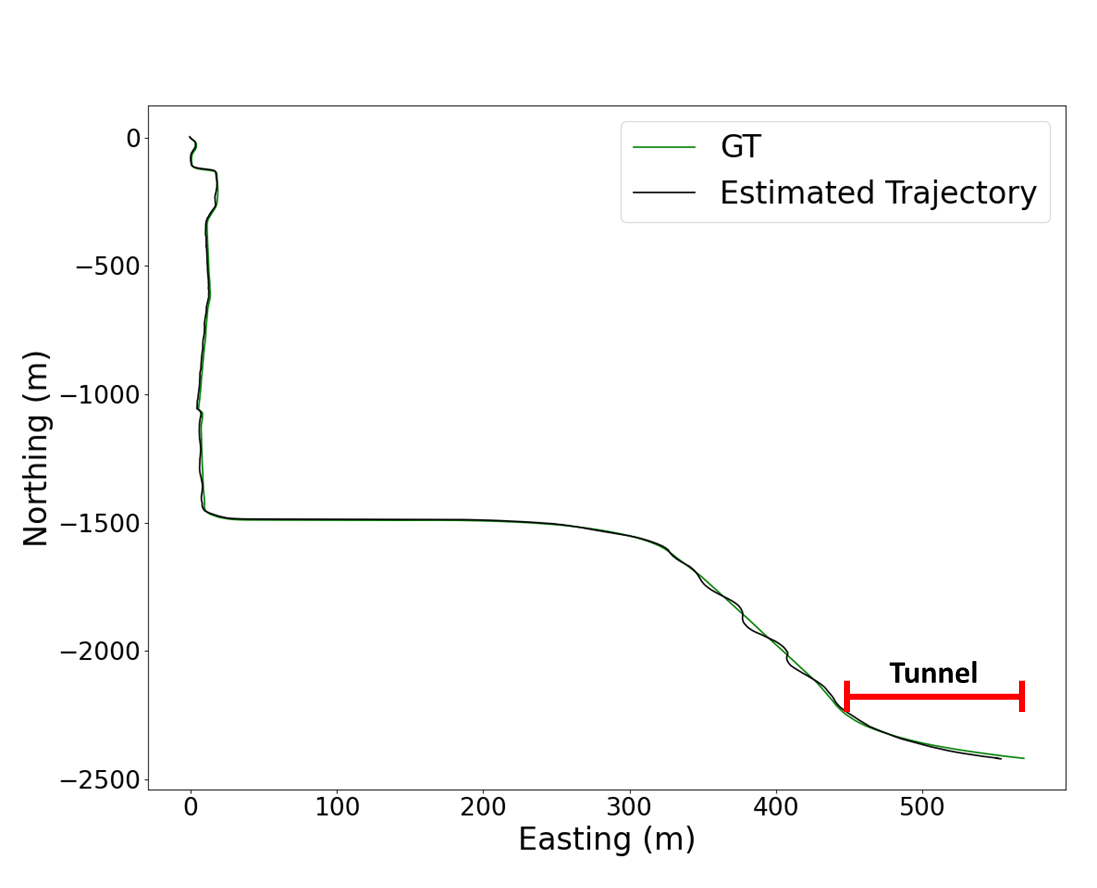} \\
        \scriptsize{ORB-SLAM3} \hspace{0.12\textwidth} \scriptsize{ORB-SLAM3}\\
        \scriptsize{\hphantom{.}} \hspace{0.21\textwidth}
        \scriptsize{+ Map-Fusion} \\
    \caption{Output trajectories of ORB-SLAM3 (left) and ORB-SLAM3 + Map-Fusion (right) for 4Seasons sequence city\_loop\_1\_train}
    \label{4seasons_correction}
\end{figure}

The qualitative results are presented in Fig. \ref{qualitative_correction} and Fig. \ref{4seasons_correction} for DROID-SLAM VO and ORB-SLAM3 VIO approaches respectively.
Fig. \ref{qualitative_correction} shows qualitatively the comparison with and without Map-Fusion for the sequences from the Oxford Robotcar Dataset. The significant drift caused by the rain is vastly reduced with the integration of Map-Fusion. 
Similarly in the 4Seasons Dataset, although the sequence city\_loop\_1\_train is marked as delocalized, Map-Fusion was able to significantly correct odometry estimates in both rain conditions and tunnel segments as shown in Fig. \ref{4seasons_correction}. 

\subsection{Real-World Results}

\begin{figure*}[h]
    \centering
    \fontfamily{cmr}\selectfont
        \includegraphics[width=0.95\textwidth]{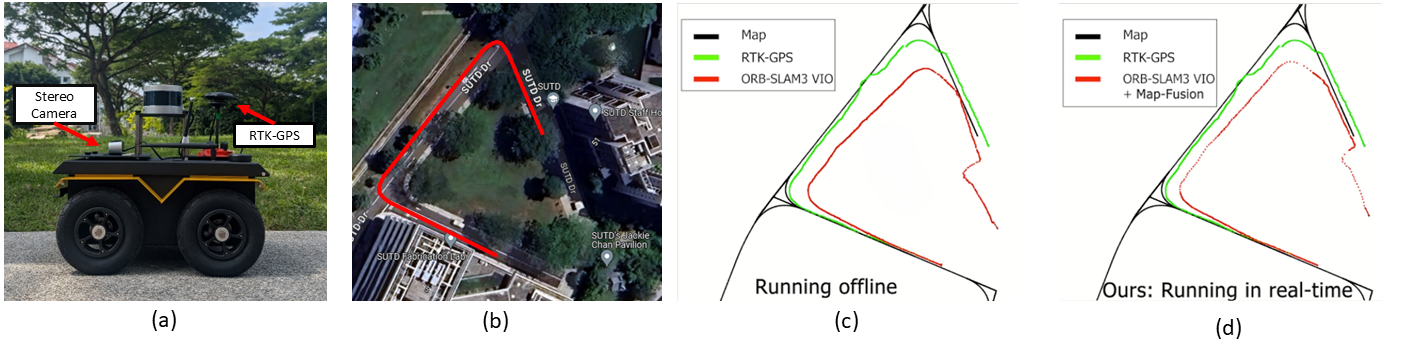} \\
    \caption{Setup of the mobile robot in real-world experiments (a). Overview of the traversed route (b). Qualitative results of running ORB-SLAM3 VIO offline (c). Qualitative results of Map-Fusion + ORB-SLAM3 VIO in real-time (d).}
    \label{real_world_results}
\end{figure*}

We also validate our proposed algorithm in a real-world environment and in real-time on a Clearpath Jackal robot. The setup is shown in Fig. \ref{real_world_results}(a) and the traversed route is shown in Fig. \ref{real_world_results}(b). 
Due to the limited CPU capacity, we reduced image size and frame rate to trade localization accuracy for real-time localization. The qualitative results for both ORB-SLAM3 VIO and Map-Fusion are shown in Fig. \ref{real_world_results}(c) and Fig. \ref{real_world_results}(d). The main source of error came from an inaccurate scale estimation likely caused by the reduced image size. We also show the more qualitative results in this link\footnote{https://youtu.be/zYXG5uA5hQU}. 
Map-Fusion achieved $2.46m$ error in clear weather and $6.05m$ error in rain weather for a $150m$ route. 
This set of experiments show that Map-Fusion could run in real-time on a hardware constrained vehicle and achieve decent localization accuracy in both clear and rain weather.

\section{Conclusion}
In this paper, we introduce Map-Fusion, a sensor-fusion based localization approach which uses map information for correcting odometry estimates in adverse scenarios such as rain conditions and intermittent GPS measurements. Map-Fusion is an affordable and practical approach that improves robustness of tracking vehicle position in urban environments. A comprehensive evaluation across $4$ different datasets with varying weather conditions showed that Map-Fusion significantly improves localization estimates in rain while also slightly improving localization in clear weather. Although Map-Fusion has limitations in being unable to correct drifts in the direction of motion and is susceptible to map errors, it could be compensated with additional sensors. Map-Fusion is useful for improving robustness of existing tracking systems and has potential applications in real-time fleet management and traffic monitoring.
\section*{Acknowledgment}
This research is supported by the Ministry of Education,
Singapore, under its Academic Research Fund Tier 2 MOE-
T2EP50121-0022.
\bibliographystyle{unsrt}
\bibliography{ITSC24}

\end{document}